\documentclass[10pt,twocolumn,letterpaper]{article}

\usepackage{wacv}
\usepackage{times}
\usepackage{epsfig}
\usepackage{graphicx}
\usepackage{amsmath}
\usepackage{amssymb}
\usepackage{booktabs}
\usepackage[accsupp]{axessibility}

\def\wacvPaperID{1375} 

\wacvapplicationstrack 

\wacvfinalcopy 

\ifwacvfinal
\usepackage[breaklinks=true,bookmarks=false]{hyperref}
\else
\usepackage[pagebackref=true,breaklinks=true,colorlinks,bookmarks=false]{hyperref}
\fi

\begin{document}


\title{Multimodal Vision Transformers with Forced Attention for Behavior Analysis}

\author{
Tanay Agrawal\\
INRIA\\
Valbonne, France
\and
Michal Balazia\\
INRIA\\
Valbonne, France
\and
Philipp Müller\\
DFKI\\
Saarbrücken, Germany
\and
François Brémond\\
INRIA\\
Valbonne, France
\and
{\tt\small\url{tanay.agrawal@inria.fr}\qquad\url{https://github.com/Parapompadoo/FAt-Transformers}}
}

\maketitle
\thispagestyle{empty}

\begin{abstract}
    Human behavior understanding requires looking at minute details in the large context of a scene containing multiple input modalities. It is necessary as it allows the design of more human-like machines. While transformer approaches have shown great improvements, they face multiple challenges such as lack of data or background noise. To tackle these, we introduce the Forced Attention~(FAt) Transformer which utilize forced attention with a modified backbone for input encoding and a use of additional inputs. In addition to improving the performance on different tasks and inputs, the modification requires less time and memory resources. We provide a model for a generalised feature extraction for tasks concerning social signals and behavior analysis. Our focus is on understanding behavior in videos where people are interacting with each other or talking into the camera which simulates the first person point of view in social interaction. FAt Transformers are applied to two downstream tasks: personality recognition and body language recognition. We achieve state-of-the-art results for Udiva v0.5, First Impressions v2 and MPII Group Interaction datasets. We further provide an extensive ablation study of the proposed architecture.
\end{abstract}


\section{Introduction}
\label{s1}

Human social behavior provides a wealth of information.
For example, facial expressions are directly linked to emotions~\cite{rosenberg2020face}, and the pattern of eye contact in a group discussion has been shown to be indicative of leadership roles~\cite{capozzi2019tracking,muller2019emergent}.
Even the highly abstract concept of personality has been shown to be related to body pose~\cite{naumann2009personality}, gaze~\cite{hoppe2018eye,larsen1996gaze}, and speech behavior~\cite{ramsay1968speech}.
With the goal to create machines that are able to interact more naturally with humans, significant efforts have been made to develop approaches that are able to sense and interpret human behavior across a wide range of scenarios and tasks~\cite{beyan2017prediction,beyan2019personality,liu2021imigue,muller2021multimediate,vinciarelli2014survey}.

A major challenge for human social behavior analysis approaches is the large variability in human behavior.
While a person's personality does play a role in their observable behavior, many other aspects such as turn-taking~\cite{birmingham2021group}, leadership~\cite{beyan2017prediction}, or rapport~\cite{muller2018detecting} are likewise major influences and the contributions of different factors are difficult to disentangle.
This issue is exacerbated by the small scale of available datasets.
Especially modern transformer-based architectures that were applied successfully to a variety of tasks~\cite{vatt,video_swin,NIPS2017_3f5ee243} struggle with such small datasets.

\begin{figure*}
    \centering
    \includegraphics[width=0.85\textwidth]{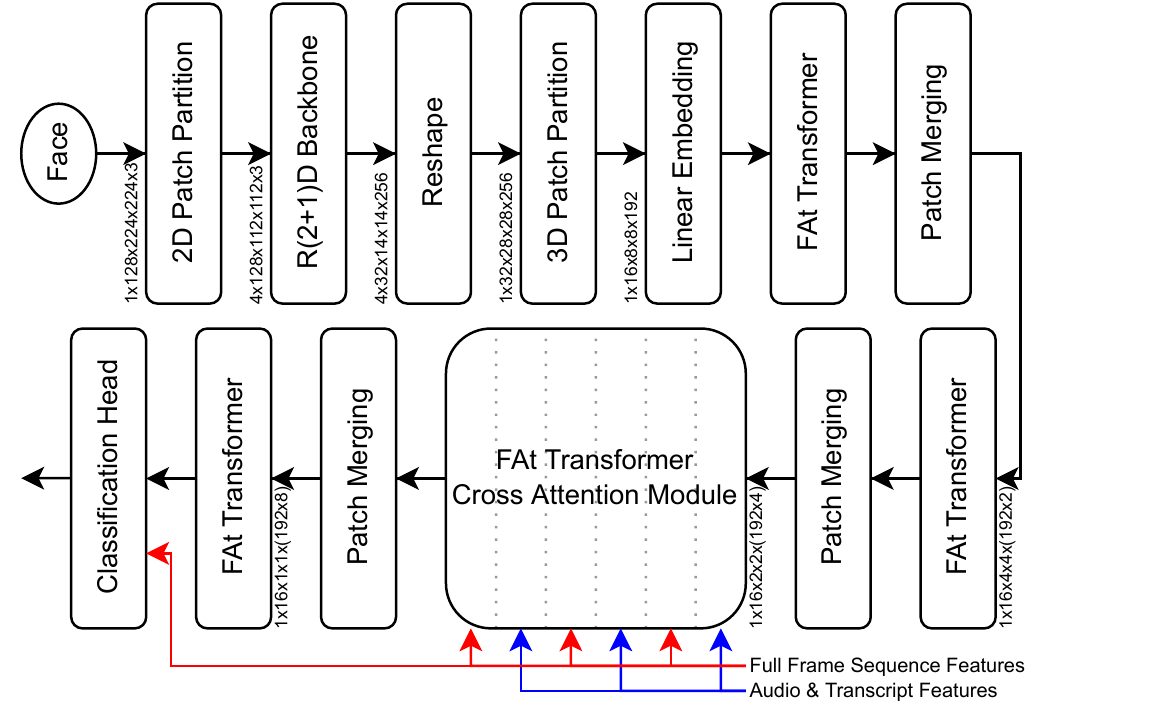}
    \caption{Face branch of the model showing the contributions of our work. Overall architecture is provided in the supplementary material.}
    \label{f1}
\end{figure*}

In this work, we introduce the novel Forced Attention~(FAt) transformer which is fit for the unique challenges present in human behavior sensing and analysis. Figure~\ref{f1} shows the architecture of the main branch.
The FAt transformer addresses the problem of large behavior variability directly by attending to important parts and reducing noise in the input. The small dataset problem is addressed implicitly as a result of faster convergence.
In detail, we introduce three distinct improvements to transformer-based human behavior analysis architectures.

First, we introduce a novel forced attention mechanism that is able to focus the processing on the relevant part of the input. 
Social interaction videos usually contain a single person who is relevant to the output task interacting with someone or something.
The remaining part is background which contains potentially misleading information.
We provide the spatial localization of the target person via a segmentation map to the network, thereby forcing the network to not attend to the background. Since the background might have important information, we observe that the network learn to assign attention to parts in the background that are also relevant to the provided background.

Second, we introduce a 2D patch partition layer in our model which combines the advantages of transformers with the robustness of convolutional layers.
We observe that training models with fully-transformers based architectures results in difficulties, such as hard to converge, due to sensitivity to background noise and the requirement of a large training sample set.
Instead, we break the videos into chunks and extract features per chunk using convolutional layers which makes the network less susceptible to pick up on noise~(non-aligned frames and random transformations such as stretching) as the convolution operation is known to be more transformation invariant as compared to attention. Extracting features from chunks allows the input to retain its spatial structure as features are only extracted from local patches without changing their arrangement.

Finally, integration of multimodal data is crucial for complex human behavior analysis tasks.
Cross-attention applied on the feature level in transformers was shown to offer effective multimodal integration for several tasks including emotion recognition~\cite{delbrouck-etal-2020-transformer}, personality recognition~\cite{palmero2021context}, and multi-view video recognition~\cite{Multiview}.
We introduce a novel variant of 
cross-attention in transformers which provides an optimised way to add multiple secondary inputs to one attention module.

We evaluate our model on two different human behavior analysis tasks: personality recognition~(high level analysis) and body language recognition~(low level analysis).
We choose personality recognition, as it is a key task in social signal processing~\cite{vinciarelli2014survey} that exemplifies the challenges faced in social behavior analysis: a non-trivial association between behavior and ground truth with random influences along with small sizes of available datasets~\cite{8999746,palmero2021context}.
Personality of a person is represented as a point in the so-called OCEAN space defined by the following five axes: O: Openness, C: Conscientiousness, E: Extroversion, A: Agreeableness, and N: Neuroticism.
In addition, we evaluate our model on body language recognition, i.e. the recognition of classes of behavior such as "fumbling", "gesturing", or "face touching".
In contrast to common action recognition tasks, body language recognition has a larger stochastic component as a result of more subjective annotation.
We achieve state-of-the-art results on three realistic human interaction datasets.
These include Udiva v0.5~\cite{palmero2021context} 
and First Impressions v2~\cite{8999746} for personality recognition, and MPII Group Interaction~(MPIGI)~\cite{acm22} for action recognition. 
We provide extensive ablation experiments to evaluate the importance of our contributions.

To summarize, the field of behavior analysis in group interactions has multiple tasks. Previous approaches work well for a particular task, adhering to its intricacies. Our aim is to introduce a generalised feature extractor which can be easily modified for a chosen downstream task. Our contributions are the following:
\begin{enumerate}
\item We introduce the Forced Attention~(FAt) mechanism which utilises segmentation maps to focus on relevant information in the input. The novelty is not in its computation, but in the way the segmentation map is incorporated in the attention mechanism of the transformer.
\item We propose an addition to Video Swin transformers. We extract features from patches of the input using a CNN based backbone making the model more robust.
\item We introduce a novel cross-attention module consisting of one main modality along with multiple other modalities.
\end{enumerate}

\section{Related Work}
\label{s2}

In the field of computer vision, CNNs have long shown to work well and have been used as backbone architectures. For tasks pertaining to videos, 3D models~\cite{DBLP:journals/corr/CarreiraZ17,tran2015learning,DBLP:journals/corr/abs-1711-11248, Qiu_2017_ICCV, DBLP:journals/corr/abs-1712-04851} have shown to give good results. But these approaches are limited by the small size of kernels with respect to the input. This is answered by vision transformers, which have a bigger receptive field with fewer parameters and have shown to have superior performance recently. Number of parameters is important for multimodal approaches as already the input requires a lot of memory. Vision Transformer~(ViT)~\cite{DBLP:journals/corr/abs-2010-11929} initialised the leaning of the community towards transformer based approaches. 
Video Swin Transformer~\cite{video_swin}, our baseline model, uses spatio-temporal local attention with spatio-temporal locality bias which has shown to perform well on various video related tasks. But its training is hard as it requires a large amount of data for converging which is presently not available for the tasks we are tackling. 
Owing to the specificity of our domain, there are some intricacies that we can exploit. Previous works, including our baseline, are not modified for these.

Since all input videos in this domain have interlocutors sitting in roughly fixed places in the frame and the cameras are also fixed, we utilise the best of CNNs and transformers to alleviate the problem of lack of data by using a CNN backbone on large local patches~(28x28 or 112x112) of the input. The patches generally have similar information across time which is easy for CNN filters to learn on finetuning. The drawback attributed to CNNs of having a small receptive field is answered as the input itself is being broken into parts and fed in parallel. It also makes for a more robust model as CNNs converge faster and are known to work better for noisy data, thus providing a cleaner input to the transformers. The transformer based part applies attention over the concatenated output of this backbone providing embeddings for higher level features for behavior analysis. This is different from regular approaches which have a CNN backbone for the whole input followed by transformers. 

As there is a lot of background noise in our tasks, we use foreground segmentation maps to provide the network with the information about which parts of the input to focus on. Since background is not to be completely removed from the input, providing the network with this information is not a trivial task. Previous works introduce a custom positional encoding for transformers which has a significant impact on performance~\cite{DBLP:journals/corr/abs-2002-12804,lajnelje,oosjdks,DBLP:journals/corr/abs-2004-03249,JMLR:v21:20-074}. But there are other stages where this information could be added. We study different configurations to pass this information to the network and show a better method for adding information to transformers for our use case, which hopefully inspires others to do similarly.

The task of combining audio and text modalities with video is challenging as they are inherently very different from each other. VATT~\cite{vatt} uses early fusion, where they input everything together. Although the earlier the fusion, the better the results, there is a trade-off with the amount of data required for training as it is harder for models with early stage fusion to converge which leads to tedious self-supervised learning. Some works design a specialised architecture for fusion at feature level~\cite{Visapp_paper,palmero2021context}. These work better but there are limitations as the fusion is done after downsampling the input features which leads to loss of information and poor cross-modality relations.~\cite{delbrouck-etal-2020-transformer,DBLP:journals/corr/abs-2003-01043,multilogue} have feature level fusion with minimal downsampling, but lack in handling specific modalities differently.
To answer this, we introduce an architecture where each branch can benefit from separate pre-training and benefit from each other with feature level fusion using a custom cross-attention module. MViT~\cite{DBLP:journals/corr/abs-2104-11227} is a multi-scale vision transformer for video recognition trained from scratch that reduces computation by pooling attention for spatiotemporal modeling. We take their method of cross-attention as inspiration and extend it to be used with multimodal transformers. Dyadformer~\cite{dyad}, which is the previous state-of-the-art for the UDIVA v0.5 dataset, uses only two modalities in each branch which leads to loss of information present in the relations among the secondary inputs. Our proposed cross attention module allows to use all modalities to be incorporated into the main branch together while including these relations.

In the next section we explain the methodology for implementing each of the above mentioned contributions.

\section{Proposed Methodology}
\label{s3}

\begin{figure*}[hbt!]
    \centering
    \includegraphics[width=0.8\textwidth]{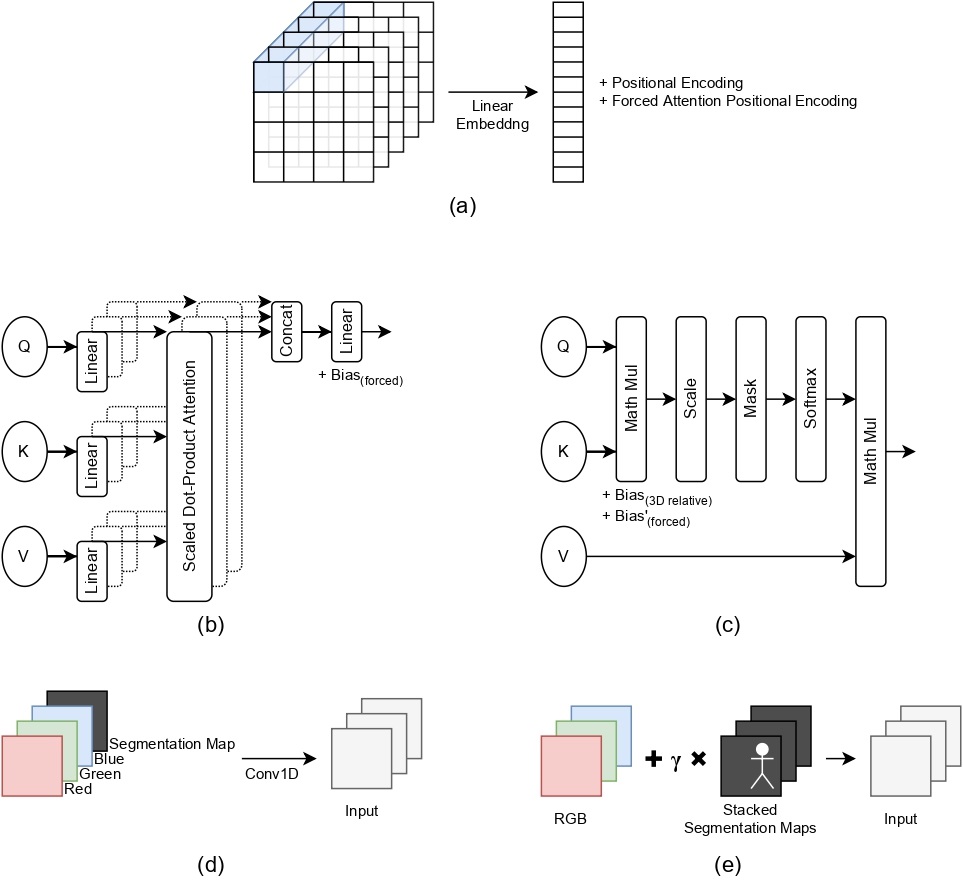}
    \caption{Different stages of adding segmentation map for forced attention in transformer encoder. (a)~shows addition of an additional positional encoding to the input with the original. (b)~shows addition of a bias to the last linear layer of multi-head self attention module. (c)~shows adding bias similar to 3D relative bias as in~\cite{DBLP:journals/corr/abs-2002-12804,lajnelje,oosjdks,JMLR:v21:20-074}. (d)~shows segmentation map being concatenated as an additional channel to raw input and then being reduced back to original shape using Conv1D. (e)~shows addition of segmentation map to each channel of the input.}
    \label{f2}
\end{figure*}

\subsection{Input to the Model and the Different Branches}
There are multiple input modalities and each of them have their own branches for processing before they are combined together using cross-attention which is discussed in Section~\ref{ss33}. For the body language recognition task, the inputs are face crop sequence, full frame sequence of the target person, and audio of the conversation. For the personality recognition task, we also have full frame sequence of the interlocutor and transcript of the target person.

Face crop sequence is treated as the main input and the other modalities are merged into it. This is because it has been established that face crops have the most relevant information for affective computing and we show that it is extendable to group interaction behavior analysis. We extract face crop coordinates using OpenFace~\cite{openface} and take crops from the original full frame. This leads to some problems as the position of the face in subsequent faces may be different and after resizing and combining back to make a video, there is a lot of distortion as discussed in Section~\ref{s1}. This urged us to devise a way to handle them. The distortions in this branch include stretching and translation as the face crops are of different resolutions and they have to be resized to the input size of 224x224. We do not use aligned face crops from OpenFace to reduce dependence on other algorithms. We break the input into 112x112 patches and pass them through a 3d convolutional backbone, R(2+1)D~(with some layers removed) and concatenate the output back again. The rest of the branch is based on Video Swin Transformers~\cite{video_swin} with our contributions added to it and is discussed later.

The full frame sequence branch has similar processing as the face branch: it is broken into 64 chunks and passed through the same convolutional backbone with shared weights. The rest of the branch is Video Swin Transformer~T~\cite{video_swin}.

The audio is passed through a pretrained model, Trill-Distilled~\cite{trill}, which is not finetuned. The obtained embedding is used as input to the cross-attention module discussed in Section~\ref{ss33}.

The transcript branch is similar to the audio one. The only difference being the model used to extract features is XLM-RoBERTa~\cite{DBLP:journals/corr/abs-1911-02116}.

\subsection{Forced Attention}
Transformers are known to be hard to train and due to the limited amount of data in the chosen domains, we choose to force the attention using foreground segmentation maps. As the background does not have relevant information, it does not have to be attended to. There are multiple ways to provide the model with this information. Adding the information encoded in the form of positional encodings is the most common way~\cite{DBLP:journals/corr/abs-2002-12804,lajnelje,oosjdks,JMLR:v21:20-074,DBLP:journals/corr/abs-2004-03249}. We study different ways and find a more suited approach to provide the model with this information. Figure~\ref{f2} shows the different techniques and they are explained below. We choose to show all methods that we tried here so that it can help others working on something similar.

\begin{figure}
    \centering
    \includegraphics[width=0.4\textwidth]{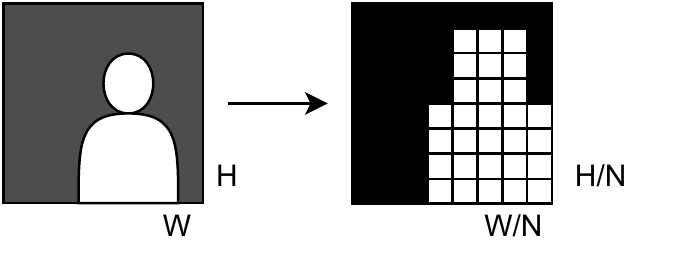}
    \caption{Visualisation of the operation to divide the segmentation map into patches to be used for input to the model in the form of the described matrix $M_1$.}
    \label{f3}
\end{figure}

The first way which is the most common is to add an additional positional encoding. It is shown in Figure~\ref{f2}(a). Transformers need positional encoding to know the location of the tokens for attention. We added another encoding to the existing ones in Swin transformers. The encoding is based on a matrix, $M_1$ of size: Nc x E, where Nc is the number of chunks~(refer to Video Swin transformers~\cite{video_swin} for details) and E is the dimension of the embedding to which the encoding has to be added. The matrix is dynamic depending on the input. Figure~\ref{f3} shows how the segmentation map is used to obtain this matrix. It is an array of ones of the size E if the chunk has pixels from the foreground or zeros otherwise. This matrix is by a weight array, $W_1$ to get the desired shape~(E in this case) and added to the input.
\begin{equation}
\begin{split}
    X = X + \mathit{SinusoidalPosEncoding} + \\
       \mathit{ForcedAttentionPosEncoding}
\end{split}
\end{equation}
\begin{equation}
       \mathit{ForcedAttentionPosEncoding} = W_1 * \mathcal{X}(M_1)
\end{equation}
$\mathcal{X}$ is a sampler function that returns an embedding of size E from the matrix $M_1$ corresponding to the chunk that is needed. Figure~\ref{f3} shows how the segmentation map is broken into parts and this function is needed to extract the required chunk for equation 2.

The second way we tried was to add a bias to the last linear layer in an attention module. This is shown in Figure~\ref{f2}(b). The bias is multiplied by one if the chunk has foreground pixels, otherwise by zero. If the bias is termed as $B_\mathit{forced}$, we can write the linear layer's output as
\begin{equation}
\begin{split}
    \mathit{Output} = \mathit{Linear}(\mathit{softmax}(Q*k^T)/\sqrt{d})*V \\
    + X) + B_\mathit{forced}
\end{split}
\end{equation}

\begin{equation}
       B_\mathit{forced} = \mathit{LearnedBias} * W_2*\mathcal{X}(M_1)
\end{equation}
Since attention is applied on input patches and encoders have the same input and output size, the same matrix can be utilised as the first approach. $W_2$ is used to change the size of chosen embedding from $M_1$ to the desired shape when required in later layers.

The third way was to add the segmentation mask as another channel in the input and use a 1d convolution layer to reduce the number of input channels back to three. This method is straight forward and does not require much explanation. It is shown in Figure~\ref{f2}(d).

In the fourth approach that we took, we add the segmentation mask to the input multiplied by a learned parameter. It is shown in Figure~\ref{f2}(e).
\begin{equation}
       X = (X * \gamma) + \mathit{PositionalEncodings}
\end{equation}

In the fifth, we add another bias similar to the first one but in the attention computation similar to 3D Relative Position Bias as in Video Swin Transformers~\cite{video_swin}. It is shown in Figure~\ref{f2}(c).
\begin{equation}
\begin{split}
    \mathit{Attention}(Q,K,V) = \mathit{softmax}(Q*k^T + \\
    B_\mathit{Relative} + B^{\prime}_\mathit{forced}/\sqrt{d})*V
\end{split}
\end{equation}
$B^{\prime}_\mathit{forced}$ is calculated in a similar way to $B_\mathit{forced}$. The only difference is that the shape is different and that is taken care by a linear layer.

The second approach gave the best results and is the one used for providing the results in the next section. We hypothesise that it is because it does not interfere with the local feature extraction from the patches but provides implicit global attention which materialises during the downsampling of channels during patch merging.

\subsection{Cross-Attention Module}
\label{ss33}

\begin{figure*}
    \centering
    \includegraphics[width=0.9\textwidth]{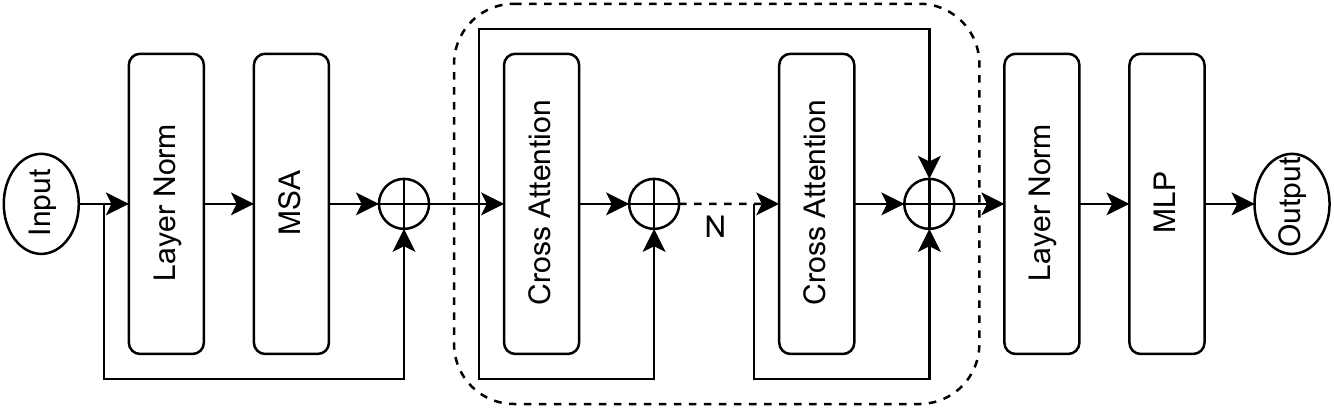}
    \caption{Cross-attention for multiple side inputs.}
    \label{f4}
\end{figure*}

This is the module that allows us to augment the face crop branch with information from the other modalities. Cross-attention has gained popularity recently and has shown to work well but there are multiple ways of implementing it. Yan et al.~\cite{Multiview} have a good comparison for some techniques and we utilise a modification of the approach that worked best for them and we extend it to use with multiple modalities to be added in one module.

In the third attention block of the face branch, we merge the modalities together. As shown in Figure~\ref{f4}, there are sequential cross-attention layers with full frame sequence and audio~(and also transcript for personality recognition). We use 1d convolution for making the channel dimension of audio and transcript features equal to 768 which is the number of channels in the third attention block for the face crops sequence branch.

Cross attention is defined as the same as is the norm, we take the query as the main branch and the key and value as the side input. 

For cross-attention with full frame sequence features, there is an additional cross attention layer after self-attention and also a residual connection imitating the self-attention layer. The parameters of this operation are zero-initialised as this helps in using pretrained weights which is the common practice~\cite{vivit}. For Udiva v0.5, there is another full frame sequence branch for the other interlocutor whose features added after the target person's as shown in Figure~\ref{f4}.

For cross attention with audio, the same approach is followed. With transcript and audio, two cross attention modules are added in place of one and both have recurrent connections along with another around both -- the same as when two full frame sequences are present.

For faster processing, we use performer~\cite{performer} in place of traditional transformer attention. This might have lead to some performance loss, but it still serves as proof of concept and allows for more experiments with the available resources and time.

\subsection{Classification Head}
The outputs of the face crop and full frame sequence branches are of the shape $B \times C \times D \times H \times W$, where $B$ is the batch size, $C$ is the number of channels, $D$ is the depth, $H$ is the height and $W$ is the width. Using an adaptive 3d pooling layer, it is reduced to $B \times C \times 1 \times 1 \times 1$ for each branch. The embeddings from each branch are concatenated along the channel dimension and then passed through a linear layer to get the output.

\section{Experiments}
\label{s4}

\subsection{Choice of Datasets}
We choose two different tasks for experiments: personality recognition~(high level) and body language recognition~(low level). Since the work is focused on group interactions, the choice of datasets is done accordingly. For personality recognition, we use UDIVA v0.5 dataset which was a challenge for ICCV 2021. This is the biggest dataset available for this task. It contains dyadic interactions involving talking and playing games. We also give results on First Impressions v2 which is also a dataset for personality recognition where people are talking facing a camera. The dataset simulates first person point of view of interactions and is a good benchmark for personality recognition as it annotates YouTube videos containing people from a diverse background and the annotations are confirmed by multiple people, so they are less influenced by personal biases. For body language recognition, we choose MPIGI dataset~\cite{acm22}. 
It has 15 body language classes, such as "fumbling", "scratching", "arms crossed", "face touching", and "grooming".
A more detailed description of the datasets is given in the supplementary material.

\subsection{Training Details}
We pretrain the full frame sequence branch on the Kinetics-400 dataset to reduce training time for the other datasets and to have a good initialisation as the other datasets are not equally big and transformers are known to be data hungry. We use the same configuration as in the Video Swin Transformer paper~\cite{video_swin}: AdamW optimiser for 30 epochs with learning rate 3e-5 for the CNN backbone and 3e-4 for the rest of the parameters. The batch size is taken as 64. 0.1 stochastic depth rate and 0.02 weight decay are used. For the CNN backbone, R(2+1)D, we use pretrained weights with the network trained on IG65-million~\cite{DBLP:journals/corr/abs-1905-00561} dataset.

For UDIVA v0.5, we use the above weights as initialisation for the full frame sequence branch. The CNN backbone is shared with the face branch. We set the learning rate for the CNN backbone and the full frame sequence branch as 3e-5 and for the remaining part of the face crop sequence branch as 3e-4. These are obtained from roughly scaling the one used in~\cite{video_swin} from the tiny variant for Kinetics-400 1K according to the batch size~(divide by square root of the ratio of batch sizes). The batch size is taken as 4. The initialisation of the face crop branch is inspired from~\cite{dhruv_paper}. The shapes do not match from the full frame sequence and we need to reshape the weights to reuse them, so we use binning methods for reshaping as described in~\cite{dhruv_paper}.

For First Impressions v2, the configuration is the same as Udiva v0.5 except the fact that there is no second interlocutor so there is only one full frame branch. We also use some additional information for this dataset that is provided as metadata. We concatenate the information with audio and transcript and pass them through a linear layer to reshape into the original size. 

For MPIGI dataset also, we do not use the information of the other interlocutor as the annotation for dyadic interaction is not available and having 3 or 4 branches for full-frame sequence is not viable due to space and time constraints. The learning rate used is double as compared to the others, i.e. 6e-4 and 6e-5 for respective parameters as explained for Udiva. Weight decay is set to 0.03. Rest of the hyperparameters remain the same. Due to imbalance of samples available for each class, we use oversampling for training samples with less frequency of occurrence and undersampling for the ones with very frequency and also a custom sampler for the dataloader which makes sure that the batches are not imbalanced.

\subsection{Comparison with the State-of-the-Art}

\subsubsection{UDIVA v0.5}
\label{sss431}

Table~\ref{t1} presents the results of state-of-the-art approaches on Udiva v0.5. Our approach greatly outperforms the previous approaches. We use the same metric that is used for the challenge, mean squared error averaged over participants. There is an interesting pattern that we saw in our experiments as compared to the others. We got better results for a few classes upon training for more epochs, but worse mean error. So, using an ensemble of trained models would give better results, but we leave that for future work. Another difference between our approach and the others is the choice of not using metadata. Other approaches use it extensively. Despite further performance improvements, we prefer a more generalised approach and so we choose to avoid the metadata, gaze and pose features. This is the reason that we hypothesise for having worse results for the Agreeableness class than hanansalam in Table~\ref{t1}. Their approach uses separate branches for male and female participants and due to data bias, it may be easy for the model to predict Agreeableness when there is a separation on its basis. But overall, we show that our model outperforms previous approaches for personality recognition in dyadic interaction videos. We do not include one paper that has the best results after us on this dataset as they do not provide results for individual classes for their best configuration. They achieve a mean error of 0.722 using the same data along with metadata about the participants which gives them a big boost. We manage to achieve a score of 0.706 with the metadata, training for only 4 epochs. The improvement is statistically significant as the 95\% confidence interval for MSE on UDIVA dataset is 0.036.

\begin{table}
\tabcolsep2pt
\begin{center}
\caption{MSE results for Udiva v0.5, as cited from the 2021 ICCV Understanding Social Behavior in Dyadic and Small Group Interactions Challenge: Automatic Self-Reported Personality Recognition Challenge \url{https://chalearnlap.cvc.uab.cat/challenge/45/track/43/result/} as of 29/08/2022.}

\label{t1}
\begin{tabular}{lllllll}
\hline\noalign{\smallskip}
Model & O & C & E & A & N & Mean\\
\noalign{\smallskip}
\hline
\noalign{\smallskip}
hanansalam & 0.711 & 0.723 & 0.867 & \textbf{0.548} & 0.997 & 0.770\\
juliojj~\cite{palmero2021context} & 0.744 & 0.794 & 0.886 & 0.653 & 1.012 & 0.818\\
f.pessanha & 0.752 & 0.687 & 0.917 & 0.671 & 1.098 & 0.825\\
gizemsogancioglu & 0.759 & 0.677 & 0.952 & 0.677 & 1.163 & 0.846\\
FAt Transformer & \textbf{0.668} & \textbf{0.624} & \textbf{0.730} & 0.590 & \textbf{0.987} & \textbf{0.720}\\
\hline
\end{tabular}
\end{center}
\end{table}
\setlength{\tabcolsep}{1.4pt}

\subsubsection{First Impressions v2}
Table~\ref{t2} shows the results of the best previous works on this dataset. For this dataset also, we greatly outperform previous work. The approach in~\cite{Visapp_paper} uses cross-attention and fusion of different modalities at the feature level, but it loses a lot of information in down sampling before cross-attention and we feel that our approach produces superior results compared to theirs because of this reason. We use the same metric for comparison that is used in other work:
\begin{equation}
\mathit{Accuracy} = 1 - {\frac{1}{N}\sum_{i=1}^{N}\left|t_i - p_i\right|}
\end{equation}
where $t_i$ are the ground truth scores and $p_i$ are the predicted scores for personality traits summed over $N$ videos.

\begin{table}
\begin{center}
\caption{Accuracy results for First Impressions v2.}
\label{t2}
\begin{tabular}{llllllll}
\hline\noalign{\smallskip}
Model & O & C & E & A & N & Mean\\
\noalign{\smallskip}
\hline
\noalign{\smallskip}
Aslan et al.~\cite{aslan2019multimodal} & 0.917 & 0.921 & 0.921 & 0.919 & 0.916 & 0.919 \\
DCC~\cite{G_l_t_rk_2016} & 0.912 & 0.911 & 0.911 & 0.916 & 0.909 & 0.912 \\
Evolgen~\cite{Subramaniam2016BimodalFI} & 0.913 & 0.914 & 0.915 & 0.916 & 0.910 & 0.914 \\
Gurpinar et al.~\cite{7899605} & 0.914 & 0.914 &0.919 & 0.914 &0.912 & 0.915 \\
PML~\cite{8014945} & 0.914 & 0.917 & 0.918 & 0.917 & 0.913 & 0.917 \\
BU-NKU~\cite{8014944} & 0.917 & 0.917 & 0.921 & 0.916 & 0.915 & 0.917 \\
Agrawal et al.~\cite{Visapp_paper}  & 0.929 & 0.926 & 0.927 & 0.929 & 0.921 & 0.926\\
FAt transformer & \textbf{0.942} & \textbf{0.951} & \textbf{0.955} & \textbf{0.949} & \textbf{0.959} &  \textbf{0.951}\\
\hline
\end{tabular}
\end{center}
\end{table}
\setlength{\tabcolsep}{1.4pt}

\subsubsection{MPII Group Interaction}
Table~\ref{t3} shows results of state of the art algorithms that we fine-tuned on MPIGI dataset~\cite{acm22}. We use accuracy and weighted average of F\textsubscript{1}-scores over the classes~(weight depending on the number of samples for each class) as metrics. F\textsubscript{1}-score for each class is defined as the harmonic mean of the precision and recall for that action class. As the dataset is heavily imbalanced, accuracy alone would not have been a good metric. We outperform even a bigger variant of the baseline that we choose, video Swin transformers~\cite{video_swin}~(we use Swin T and compare with Swin B) owing to our contributions which shows their saliency for this field. We also outperform other approaches that give state-of-the-art results on action recognition which is the closest field to this dataset with numerous previous works.

\begin{table}
\begin{center}
\caption{Accuracy and F\textsubscript{1}-score results for MPII Group Interaction.}
\label{t3}
\begin{tabular}{lll}
\hline\noalign{\smallskip}
Model & Accuracy & Weighted F\textsubscript{1}-Score\\
\noalign{\smallskip}
\hline
\noalign{\smallskip}
TSN~\cite{tsn}  & 0.443 & 0.442\\
TSM~\cite{tsm}  & 0.607 & 0.599\\
Video Swin B~\cite{video_swin} & 0.656 & 0.637\\
FAt Transformer & \textbf{0.692} & \textbf{0.685}\\
\hline
\end{tabular}
\end{center}
\end{table}
\setlength{\tabcolsep}{1.4pt}

\subsection{Ablation Study}
\begin{table}
\begin{center}
\caption{MSE results for Udiva v0.5 as ablation study.}
\label{t4}
\begin{tabular}{lllllll}
\hline\noalign{\smallskip}
Model & O & C & E & A & N & Mean\\
\noalign{\smallskip}
\hline
\noalign{\smallskip}
FAt Transformer & \textbf{0.668} & 0.624 & \textbf{0.730} & \textbf{0.590} & \textbf{0.987} & \textbf{0.720}\\
w/o Forced Attention & 0.705 & 0.704 & 0.873 & 0.966 & 1.264 & 0.902\\
w/o CNN backbone & 0.672 & \textbf{0.622} & 0.850 & 0.737 & 1.208 & 0.818\\
w/o Cross Attention & 0.813 & 0.776 & 0.794 & 0.619 & 1.118 & 0.824\\
w/o Late Fusion & 0.709 & 0.629 & 0.751 & 0.602 & 1.012 & 0.741\\
w/o Audio & 0.797 & 0.919 & 0.874 & 0.844 & 1.079 & 0.903\\
w/o Transcript & 0.839 & 0.876 & 0.832 & 0.760 & 1.150 & 0.911\\
\hline
\end{tabular}
\end{center}
\end{table}
\setlength{\tabcolsep}{1.4pt}

\subsubsection{Main contributions}
When segmentation map is absent from the input, we noticed that the model performance oscillates in every epoch. In Table~\ref{t4}, we give results with the best mean validation score across the OCEAN classes. Due to the bias in the dataset and maybe even the personality classes, we saw that in a particular epoch the model performed well either on O, C and E classes together or A and N classes. In Table~\ref{t4} for instance, it can be seen that the model does not perform well on A and N without~(w/o) forced attention for the chosen epoch. We saw the same phenomenon in a separate experiment where we tried an approach similar to VATT~\cite{vatt} without contrastive learning. So, we conclude that forced attention indeed helps the model converge.

Without patch partitioning and a CNN backbone, the transformer behaves similarly with the exception of extroversion class. This corroborates our assumption that this contribution also helps in convergence.

Using only one side input and changing it in successive blocks in place of the proposed module allows the network to converge properly, but the performance across all classes decreases. This contribution allows our network to better exploit the correlations between the modalities and this ablation study supports the claim.

\subsubsection{Late Fusion}
We try two different configurations for the classification head using only features from the face branch, and with both full-frame and face crop sequence branches concatenated along the channel dimension. The latter one gave the best results, but they are not very different from using only face features which is the second row in Table~\ref{t4}. This shows that the cross-attention module incorporates the relevant information from the other branches~(other than the face crop sequence) properly and this explains why late-fusion does not have a great impact on results.

\subsubsection{Different Modalities}
We show how the model works when audio and transcript inputs are missing. There is significant reduction in performance when either modality is missing which shows the importance of both. It can be seen that some classes are more affected by the absence of these modalities. Class A has the most reduction in performance, but this might also be explained by the fact that the model does not properly converge for this class and having more input modalities helps with that. On the other hand, results on C seems to be highly dependant on these modalities as the model is able to converge for this class in the forced attention ablation study, but not without these. These results also show that our architecture works well in merging information from audio and transcript into the face crop sequence. 


\section{Conclusion}
\label{s5}

We presented a method to efficiently incorporate multiple modalities into one model for human behavior analysis. Our model outperforms the state-of-the-art on three different datasets. Udiva v0.5 and First Impressions v2 tackle the high level problem of personality recognition while MPIGI tackles the low level problem of body language recognition. 
For the respective datasets, we achieve 0.050 lower MSE (ignoring dyadformer: described in detail in section~\ref{sss431}), 1.3 \% higher accuracy, and 3.6 \% higher accuracy. 
Through ablation studies, we showed that each of our contributions has a significant impact on the performance of the model. We studied different ways to incorporate information in a transformer and show the impact of the one with the best performance through the visualisation of attention on an example input frame. We hope that this work inspires more work in this area as it is not widely explored.

\subsection*{Acknowledgments}
\noindent This work was supported by French National Research Agency under the UCA\textsuperscript{JEDI} Investments into the Future, project number ANR-15-IDEX-01, and by German Ministry for Education and Research, grant number 01IS20075.

{\small
\bibliographystyle{ieee_fullname}
\bibliography{egbib}
}

\end{document}